# Binocular Gaze Estimation with Single Camera and Single Light Source

Tongbing Huang
7Invensun Technology Co.Ltd.
Beijing,China
huangtb@7invensun.com

Yang Fu
TnB Galaxy
Beijing,China
624933405@qq.com

Yunfei Wang
7Invensun Technology Co.Ltd.
Beijing,China
wangyf@7invensun.com

Zhaocan Wang
University of Science and Technology Beijing
Beijing,China
wzc1ws2@163.com

## ABSTRACT
According to commonly consented theories, the minimum hardware requirement for gaze tracker is one camera and two light sources to realize gaze estimation with free head movements [2]. However, in some scenarios such as eye tracking on mobile devices, it is preferable to use less components, especially light sources. We propose a gaze estimation method with one camera and one light source. A "virtual light source" is introduced, which is geometrically placed symmetrically to the real light source with respect to the camera, and generates a "virtual glint" in the acquired image. We estimate the "virtual glint" by exploiting the relationship between the distance between two pupils and two glints in the captured image, and estimate the gaze with polynomial regression assuming two light sources are available. A new normalization factor for regression method is verified, which turns out to be practical for one-glint system. The performance is proved to be acceptable, while degradation is noticed compared to system with two actual light sources.

## CCS Concepts
**• Computing methodologies~Model verification and validation**

## Keywords
Gaze estimation, Polynomial regression, Binocular, Video oculography

## 1. INTRODUCTION
Gaze estimation is a technique to tell where the user is looking, the estimated gaze is represented by either a gaze point or gaze direction. Existing methods include scleral coil method [11], electrooculography (EOG) [12], and video oculography [2-10]. Gaze estimation with video oculography has been proved to be the most practical approach for commercial and academical purposes [1]. Two common methods are geometric method [2, 3] and regression method [4, 5], both of which require a minimum of one camera and two light sources to enable head movement [2, 4]. Geometric method, which relies on 3D model of eye, estimates the line of sight by solving parameters of eye model. With one camera and two light sources, there are enough equations to solve the unknown parameters without knowing distance information of the user [2]. Regarding regression method, with one off-axis light source, noticeable mean error exists. Such drawback caused by asymmetry glint could be eliminated by using two light sources and define the pupil-cornea reflection ($PCR$) vector as the midpoint of the two glints [4]. It also points out that it is susceptible to perpendicular head movement simply using $PCR$ as regression input, but more robust with normalized $PCR$ vector. Experiments indicate that the best normalization factor is inter-glint distance. A study on mobile phone eye tracking [6] is based on one camera and two light sources embedded on the phone. The two light sources are place apart from each other diagonal to the screen because the glint in the image might adjoin each other if they are placed too close, making feature extraction erroneous. However, such design is not quite practical since it is very difficult to have both glints available due to the obstruction of eyelids.

For commercial applications, it is preferable to use fewer light sources. Take eye tracking on mobile phone for example, the manufactures prefer to save the space for front-facing components to accommodate bigger screen. According to our observations and mobile platform visibility study [7, 8], for consideration on glint(s) visibility, the light sources should not be place too far from the camera. And the most practical solution is too use only one light source.

[4] made an attempt with system bearing only one light source, the method is 3D model based which introduced two restraints to combat the loss of the second glint, 1) the cornea curvature centers distance is constant, 2) two visual axes intersect on the screen plane. The accuracy is reported to be less than 1 degree although noticeable degradation occurs compared to 3D method with two glints. The authors assume the interpupillary distance (*IPD*) is a known constant. As a matter of fact, the IPD varies between

individuals, thus, the method performs poorly on those with extraordinary IPD.

In order to decrease the number of required components and to overcome said problems, this paper proposes a 2D gaze estimation method for system with one camera and one off-axis light source. The key idea is to assume that there is a "virtual glint" created by a "virtual light source" which does not actually exist. The virtual light source is assumed to be placed symmetrically to the actual light source with respect to the camera. In practice, both light sources and the nodal point of the camera are geometrically aligned and remains roughly horizontal. According to the principal of spherical mirror imaging, in the acquired image, the "virtual glint" is horizontally apart from the actual glint by a scale referred to as inter-glint distance ($IGD$), which represents the Euclidean distance between two glints in the acquired image. Knowing the location of the actual glint and the $IGD$, the location of "virtual glint" could be estimated. An approximation is provided, which indicates that the $IGD$ and $IPD$ follow a constraint invariant to the eye-camera distance. The coefficient of such constraint function could be acquired with a calibration process. With the virtual glint, the actual glint, and the $PCR$ vector derived therefrom, gaze point could be derived using polynomial regression. It is also pointed out that the square of $IPD$ is equivalent to $IGD$ as $PCR$ normalization factor. Binocular information, i.e. image plane $IPD$ in this work, is used to calculate gaze for each eye, the gaze combining logic is not discussed herein. The proposed method is presented in Section 2. Experiments are conducted to verify the proposed method; the result is shown in Section 3. Conclusions and comments are given in Section 4.

## 2. METHOD

The proposed method is based on classic polynomial regression, which is well explained by [4]. The classic method is not elaborated here. This section focuses on the main contribution, which is the estimation of the $IGD$ and the "virtual glint".

According to eye anatomy, light reflects on boundary between the lens and cornea, creating several reflections, such image is called Purkinje image [9]. The first Purkinje image is the most evident one and is referred to as the glint. The glint is most likely to fall on the surface on the cornea, unless eye rotates a large angle away from the front. The cornea surface is modeled as a spherical mirror as shown in Figure 1, in which the cornea curvature radius is $r$. The light source $l$ creates its image $q$ behind the spherical mirror, according to the property of the spherical mirror, $q$ is roughly $r/2$ from cornea radius center $C$. The nodal point of camera, with focal length $f_{camera}$, is at point $O$, which is aligned with $l$ in the direction of $CO$. Let $D$ be the distance between $O$ and $C$, which is consisted of $d_1$ and $d_2$, where $d_2 \approx r/2$.

Let the distance from $q$ to the axis of $CO$ be $v$, and the corresponding scale on the image plane be $u$. The geometric constraint on $v$ and $u$ can be written as equation 1.

$$\frac{v}{d_1} = \frac{u}{f_{camera}} \tag{1}$$

$l$ and its image $q$ in the spherical mirror are colinear with the cornea curvature center $C$, giving us equation 2.

$$\frac{v}{d_2} = \frac{L}{D} \tag{2}$$

With (1) and (2), we obtain equation 3.

$$ud_1D = d_2Lf_{camera} \tag{3}$$

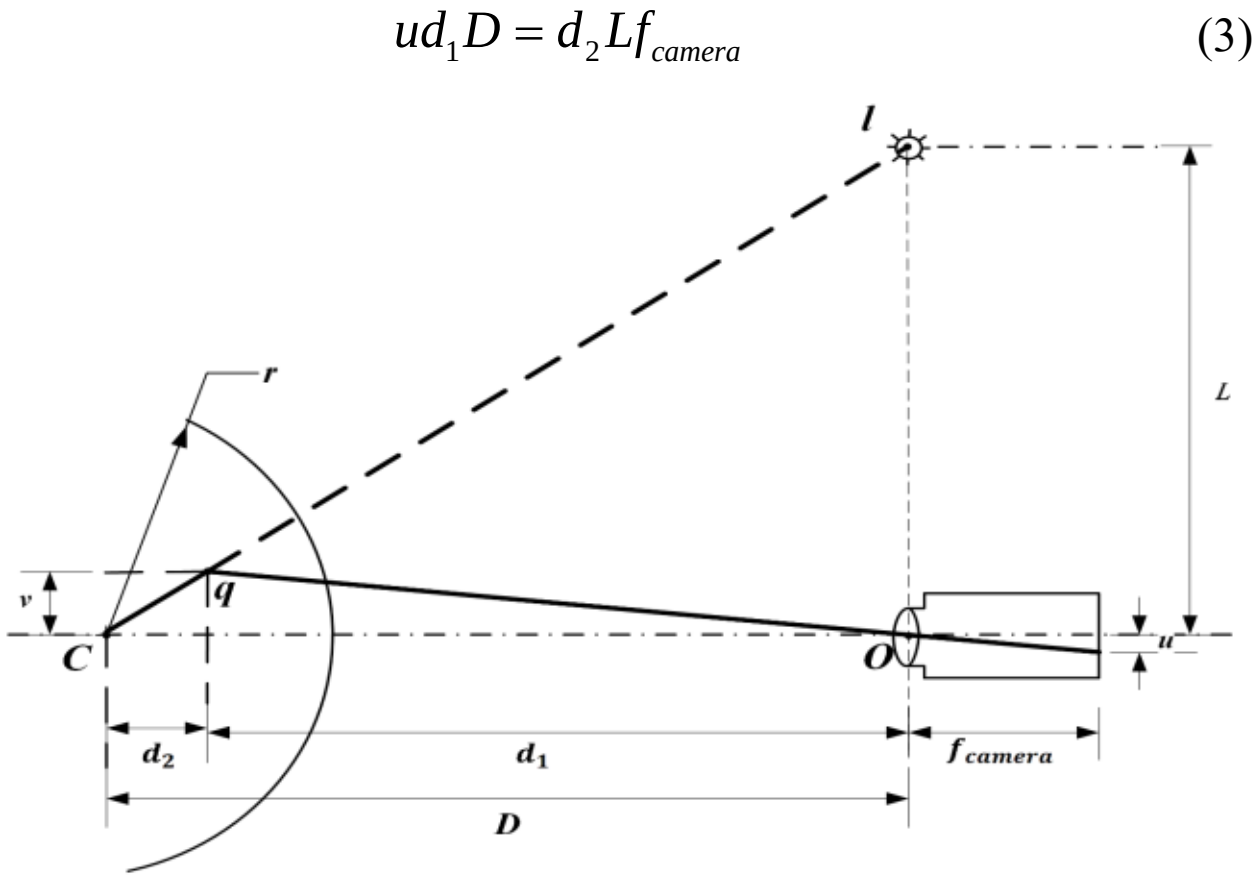

**Figure 1: 3D model of cornea surface, which is modeled as a spherical mirror.**

The radius of cornea curvature is typically 7.8mm [10], which is negligible compared with $D$, typically ranging between 400mm to 800mm. Therefore, an approximation $d_1 = D$ is applied on (3). By substituting $d_2$ with $r/2$, $d_1$ with $D$, we obtain equation 4.

$$uD^2 = rLf_{camera}/2 \tag{4}$$

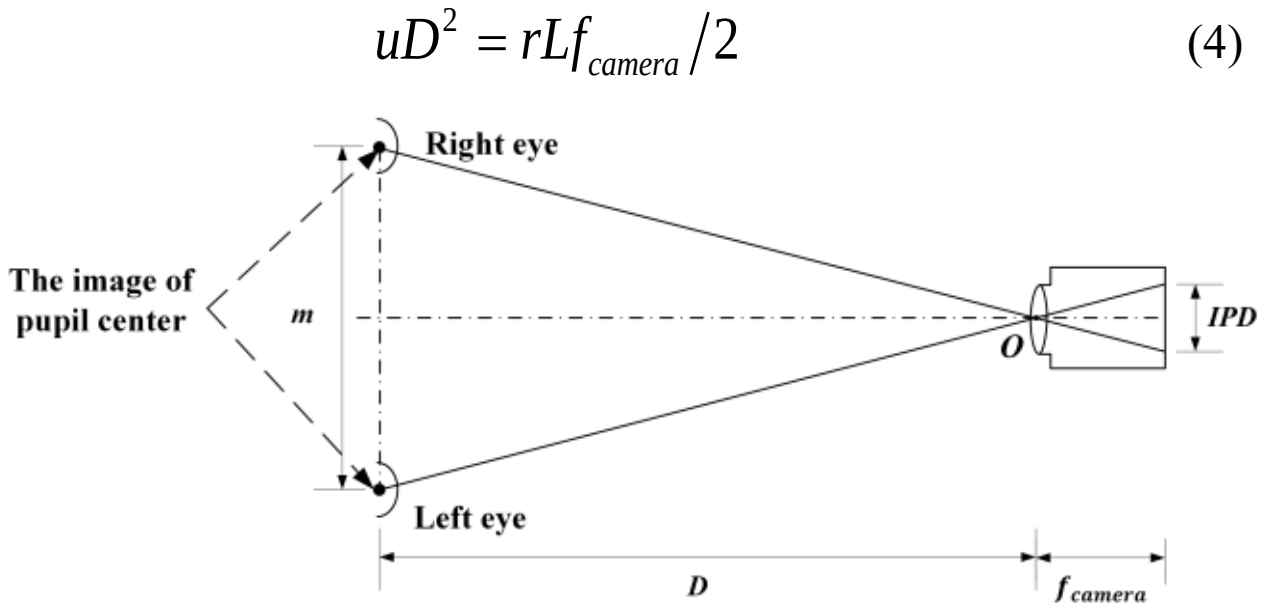

**Figure 2: Geometry of pinhole camera imaging. The image of pupil center is close to but different from the actural pupil center allowing for the refraction of optic path on cornea surface.**

As shown in Figure 2, the actual interpupillary distance $m$ between two eyes and $IPD$ on the image plane follow a geometric relationship expressed by (5) using pinhole camera model. Note that (5) holds on condition that the head yaw angle is zero. If lateral head rotation exists, such relationship won't hold unless compensation is made considering rotation angle.

$$\frac{m}{D} = \frac{IPD}{f_{camera}} \tag{5}$$

For system with symmetric light source, $IGD = 2u$. Combining (4) and (5), and replace $u$ with $IGD/2$, the relationship between $IPD$ and $IGD$ is given by equation 6.

$$\frac{IPD^2}{IGD} = \frac{2m^2}{r^2L^2} \tag{6}$$

Since the actual interpupillary distance $m$, cornea radius $r$ are personal dependent and invariable, light source-camera distance $L$ is fixed, $IPD^2/IGD$ is constant for one specific user. It should be mentioned that, the actual interpupillary distance $m$ is the distance between the image of pupil center, which, owing to the refraction by cornea, does not strictly collocate with the actual interpupillary distance. Additionally, the truly invariant physical distance is eye ball rotation center distance, however, the difference between interpupillary distance and eye ball rotation center distance is negligible. The term $IPD - IGD$ constant refers to $k = 2m^2/r^2L^2$ as shown in (6), and will be used in the following documentation.

However, $IGD$ is not available for system with one light source. In order to compute $IGD$, we assume there is one symmetric virtual light source which, accordingly, creates a virtual glint. Calibration at two different distances is required, the user is asked to look at a same calibration point in the middle of the screen while holding head straight to the camera.

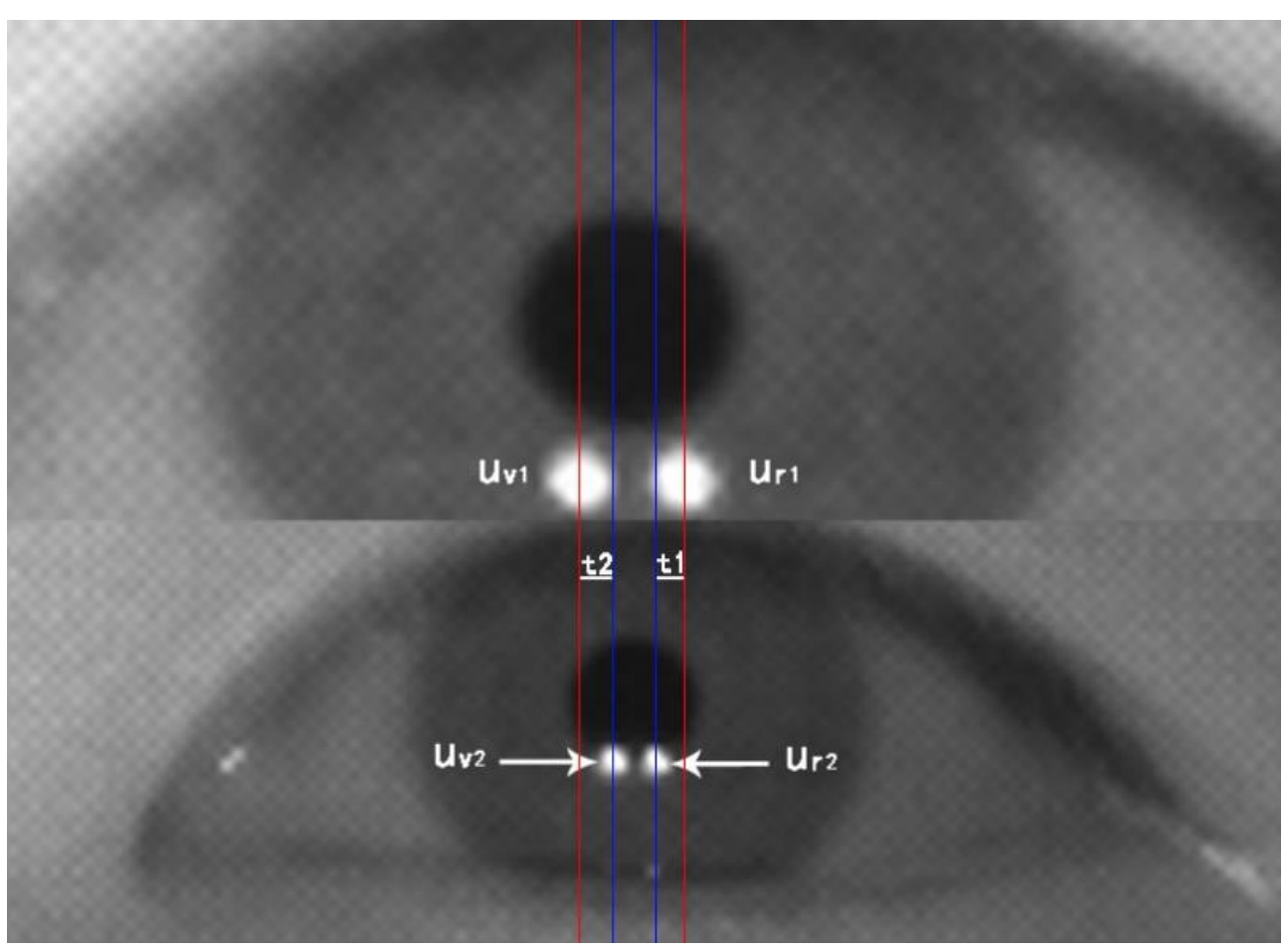

**Figure 3: Eye image captured at two different distance. $u_{r1}$, $u_{r2}$ are the location of the actual glint at the first and second distance, $u_{v1}$, $u_{v2}$ are virtual ones at the first distance and second distance. From the first distance to the second distance, they converge by a same extent if calibrated as requested**

As illustrated in Figure 2, the upper eye image is captured when the user is close to the camera, the lower one when the user is far away. At the first distance, $u_{r1}$ is the glint created by the actual light source, $u_{v1}$ regarded as the glint created by the virtual light source. Similarly, we have $u_{r2}$ and $u_{v2}$ at the second distance. Moving from the first distance to the second distance, the real glint and virtual glint converge by $t_1 = u_{r1} - u_{r2}$ and $t_2 = u_{v1} - u_{v2}$, respectively. By either using pupil center as anchor point for glints or align the pupil centers in the two images, the key basis to estimate $IGD$ is that real glint and virtual glint converge by a same extent, i.e. $t_1 = t_2$. Such assertion is based on massive observations, the analytical derivation is not to be elaborated, but will be verified via gaze estimation performance.

Based on the afore proved approximation (6), the relationship between $IPD$ and $IGD$ at two different distances is written as (7) and (8), $IGD$ converge restriction as (9). $IPD_1$ and $IPD_2$ represent $IPD$ at two distances and are known, leaving us three functions with three unknowns. Thus, the constant $k$ could be solved from such a calibration process. Knowing $k$ and $IPD$, $IGD$ could be estimated, which consequently gives us the location of the virtual glint using $IGD$ and the actual glint.

$$\frac{IPD_1^{\ 2}}{u_{r1} - u_{v1}} = k \tag{7}$$

$$\frac{IPD_2^{\ 2}}{u_{r2} - u_{v2}} = k \tag{8}$$

$$u_{r1} - u_{r2} = u_{v2} - u_{v1} \tag{9}$$

Polynomial regression method, with full second order polynomial, is used to evaluate the proposed method. $PCR$ vector is expressed as the midpoint of the real glint $u_r$ and virtual glint $u_v$, as $PCR = (u_r + u_v)/2$. The estimated $IGD$ or $IPD^2$ is used as the normalization factor.

# 3. EXPERIMENT AND RESULT

The experiment is performed on 5 test subjects, 2 sessions per test subject. The test subjects are asked to watch with nude eyes (*w/o glasses)* and with eye glasses (*w glasses)* (in the first and second session. Daily, these test subjects wear eye glasses with negative power ranging from -1.0 to -4.0, so that we could compare the impact of eye glasses on same individuals. It is reported by all the test subjects that they are able to identify calibration points with nude eyes.

For each session, the test subject is asked to perform a one-point calibration at 3 different distances. This process allows us to computer the $IPD - IGD$ coefficient. Afterwards, the test subject performs a 9-point calibration at 60cm and 70cm, the data at 60cm is used as training set to generate polynomial coefficients, that at 70cm is used as test set to evaluate the performance. The 3-by-3 points grid are rendered on a 24-inch screen, with 23cm horizontal interval and 12cm horizontal interval. At both distances, the users are allowed to move head naturally to follow different calibration points. Only perpendicular head translation is considered and evaluated in this work, while the impact of lateral translation is not discussed herein. The eye tracker used in the experiment is a prototype eye tracker provided by 7invensun [1], which is designed for desktop and laptop eye tracking. The eye tracker has one camera and two symmetric IR light sources placed 24cm apart.

During the experiment, both light sources are turned on, one of the light sources is regarded as the real light source, the other as the virtual light source. To evaluate the proposed method, we deliberately ignore the virtual light source and corresponding virtual glint, employing the real one alone. The virtual glint, although exists, is used by classic method to provide reference performance to be compared with the proposed method.

The test result on $IPD - IGD$ coefficient k is shown in Table 1. The coefficient with two real glints, at three distances, is given by $k_1, k_2, k_3$ respectively. Estimated k is estimated with one real glint using proposed method. The error of $k$ is defined as $k_{error} = |k_{est} - k_{ave}|/k_{ave}$, where $k_{ave} = (k_1 + k_2 + k_3)/3$.

First, to examine the $IPD - IGD$ coefficient $k$, for each test session, $k$ at different distance are very similar. The standard deviation for each session's $k_1,\ k_2,\ k_3$ is averaged, which is as low as 0.01898 (mean: 3.125) . Which supports the assumption that the $IPD - IGD$ coefficient could be approximated as a constant.

The $IPD - IGD$ coefficient could either be normative or determined by personal calibration. Theoretically, the coefficient is determined by the physical interpupillary distance and the cornea curvature radius, which differ between individuals. Such difference infers that personal calibration may provide better estimation. According to the data in our experiment, the estimated coefficient is more erroneous than a normative coefficient derived by averaging $k_{ave}$ of all test subjects. To assess this problem, further experiments should be done to evaluated the individual difference. Regarding the impact of eye glasses, the $IPD - IGD$ coefficient without eye glasses is 5.85% less on average. Such phenomenon is caused by eye glasses which magnify/diminish the eyes for the camera. As mentioned, all the test subjects wear eye glasses with negative power, leading to smaller $IGD$ while having much smaller impact on $IPD$, which consequently leads to larger $IPD - IGD$ coefficient.

For the ten test sessions, the estimation error of k ranges from 1% to 8.7%, with an average of 5.3%. The method is further evaluated by observing gaze accuracy.

The performance of different methods is shown in Figure.4. Angular mean error (ME) and mean square error (MSE) are calculated with test dataset at 70cm. $G_2$, $G_{1+1}$, $G_1$ represents classic method with two actual glints, proposed method with one actual glint and one estimated glint, method with only one actual glint, respectively. Normalization methods are distinguished by $N_{igd}, N_{ipd^2}, N_{ipd}, N_1$ which represents $IGD$, $IPD^2$, $IPD$ and constant 1 (no normalization applied).

Regarding normalization, as shown by the performance of classic methods $G_2N_{igd}$, $G_2N_{ipd^2}$ and $G_2N_{ipd}$, the result normalized by $IGD$ and $IPD^2$ are very close, while that normalized with $IPD$ is much worse. Such result coincides with the assertion that $IPD^2/IGD$ could be approximated with a constant, both $IGD$ and $IPD^2$ could be used to combat the impact of perpendicular head movement. Noting that normalization with $IPD^2$ is slightly better than $IGD$, this is due to the fact that $IGD$ is much shorter than $IPD$ and is more vulnerable to noise, thus the normalization with $IGD$ tends to be more erroneous. However, such result and conclusion are derived by restraining the users' head orientation. If head yaws, such conclusion becomes inaccurate unless compensation is made.

**Table 1. Estimated IPD-IGD coefficient k and its error compared with actual k at 3 different distance**

| | **Test Subjects 1** | | **Test Subjects 2** | | **Test Subjects 3** | | **Test Subjects 4** | | **Test Subjects 5** | |
|---|---|---|---|---|---|---|---|---|---|---|
| | ***w/ glasses*** | ***w/o glasses*** | ***w/ glasses*** | ***w/o glasses*** | ***w/ glasses*** | ***w/o glasses*** | ***w/ glasses*** | ***w/o glasses*** | ***w/ glasses*** | ***w/o glasses*** |
| $Distance_1 - k_1$ | 3.168 | 2.969 | 3.140 | 3.006 | 3.275 | 2.957 | 3.219 | 3.053 | 3.261 | 3.120 |
| $Distance_2 - k_2$ | 3.130 | 2.986 | 3.206 | 3.011 | 3.272 | 2.957 | 3.227 | 3.078 | 3.314 | 3.103 |
| $Distance_3 - k_3$ | 3.130 | 3.044 | 3.219 | 3.000 | 3.256 | 2.958 | 3.209 | 3.085 | 3.262 | 3.130 |
| $Estimated - k$ | 3.031 | 2.832 | 3.308 | 3.268 | 3.502 | 3.156 | 3.434 | 3.041 | 3.384 | 2.924 |
| $k_{error}$ | 3.6% | 5.6% | 3.8% | 8.7% | 7.2% | 6.7% | 6.7% | 1.0% | 3.2% | 6.2% |

[1] Beijing 7invensun Technology Co., Ltd. (www.7invensun.com) is a company focusing on eye tracking technologies and applications. The prototype eye tracker used in this work is not yet released.

For methods only depending on the actual glint, $G_1N_1$ is most affected by head movement, while $G_1N_{ipd^2}$ is much better with normalization using $IPD^2$. Comparing $G_1N_{ipd^2}$ and $G_2N_{ipd^2}$, the method with two actual glints evidently prevails, such advantage is caused by the symmetry of the glints, which allow us to compute $PCR$ by averaging two glints. However, such symmetry is inaccessible for $G_1N_{ipd^2}$, which is consequently vulnerable to head movements.

For the proposed method, normalization with $IPD^2$ and $IGD$ are essentially equivalent since $IGD$ is calculated by $IGD = IPD^2/k$. Therefore, only $G_{1+1}N_{ipd^2}$ is shown in the Table. With same normalization technique, the proposed method is better than $G_1N_{ipd^2}$ which depends only on the actual glint, but worse than $G_2N_{ipd^2}$ which depends on two actual glints. Ideally, the proposed method should have equal performance to classic method since the virtual glints is estimated. However, affected by the estimation error on the virtual glint, the proposed method is not as good as the classic method. Such estimation error is mainly contributed by the assumption that left and right glint converges to a same extent at two different distances, which leads to inaccurate estimation of $IPD - IGD$ coefficient $k$.

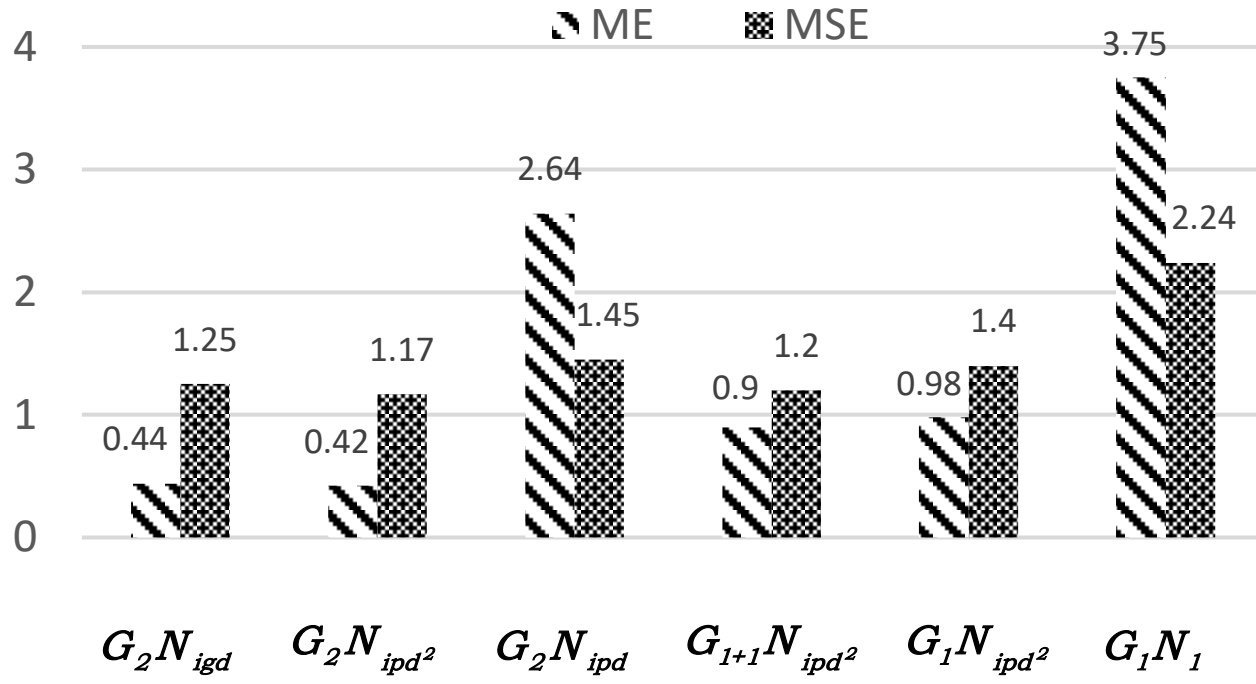

**Figure 4: ME and MSE in angle for different methods.**

## 4. CONCLUSIONS

This work proposes a new gaze estimation method for system with one camera and one light source. With such setup, the estimated gaze is inaccurate unless the eye-camera distance is known. The proposed method exploits the binocular information, i.e. $IPD$, and analytically points out that $IPD^2/IGD$ could be approximated as a constant. The virtual glint is estimated with $IGD$ and known real glint. The gaze is estimated with polynomial regression using $PCR$ represented by the midpoint of two glints and normalized by $IGD$.

The proposed method is verified with experiments. Statistics indicate that the $IPD - IGD$ coefficient $k$ varies between individuals but remains almost invariant on a same individual. The performance is further evaluated by observe gaze accuracy, which infers that on condition that there is only one light source, the proposed method bears worse ME, while the overall performance is acceptable.

This work is based on one off-axis light source. One alternative is to use one on-axis light source. With on-axis light source, $PCR$ is naturally immune to aforementioned asymmetry impact, consequently leads to better gaze estimation accuracy. Since $IGD$ is not available for one-glint system, using $IPD^2$ as normalization factor could further enhance the performance.

As for eye tracking on mobile phone, the light source might not be placed 12cm away from the camera, but much closer. It could be expected that if the light source is placed closer to the camera the asymmetry impact becomes less significant, making the performance of $G_{1+1}N_{ipd^2}$ and $G_1N_{ipd^2}$ more comparable. In practice, estimation of the virtual glint might have insignificant contribution while using $IPD^2$ as normalization factor may significantly combat perpendicular head movement.